\documentclass[10pt,twocolumn,letterpaper]{article}

\usepackage[pagenumbers]{iccv} %

\usepackage[utf8]{inputenc} %
\usepackage[T1]{fontenc}    %
\usepackage{url}            %
\usepackage{booktabs}       %
\usepackage{amsfonts}       %
\usepackage{nicefrac}       %
\usepackage{microtype}      %
\usepackage{colortbl} %
\usepackage{comment}
\usepackage{float}
\usepackage{colortbl}
\usepackage{siunitx}
\usepackage[table,dvipsnames]{xcolor}
\usepackage[section]{placeins}

\definecolor{table_shade}{RGB}{200, 250, 200} %

\usepackage{multirow}

\definecolor{light}{rgb}{0.5, 0.5, 0.5}
\def\light#1{{\color{light}#1}}

\usepackage{graphicx}
\usepackage{subcaption}

\definecolor{tabg}{rgb}{0.0, 0.8, 0.6}
\definecolor{tabr}{rgb}{1.0, 0.0, 0.22}

\usepackage{tcolorbox}
\newtcolorbox{formattedquote}{
    colback=blue!3!white,
    colframe=blue!20!white,
    fontupper=\ttfamily\footnotesize,
    boxsep=-5pt %
}

\usepackage{adjustbox}
\usepackage{calc}

\definecolor{iccvblue}{rgb}{0.21,0.49,0.74}
\usepackage[pagebackref,breaklinks,colorlinks,allcolors=iccvblue]{hyperref}

\def\paperID{13963} %
\def\confName{ICCV}
\def\confYear{2025}

\title{\datasetName: A Challenging \underline{GR}aph \underline{A}nalysis \underline{B}enchmark for Large Multimodal Models}

\author{
  \textbf{Jonathan Roberts\textsuperscript{1}},
  \textbf{Kai Han\textsuperscript{2}},
  \textbf{Samuel Albanie}
\\
  \textsuperscript{1}University of Cambridge,
  \textsuperscript{2}The University of Hong Kong
\\
  \tt\small{jdr53@cam.ac.uk, kaihanx@hku.hk, samuel.albanie.academic@gmail.com} \\
\small\url{https://grab-benchmark.github.io}
  }

\newcommand{\datasetName}{GRAB\xspace}
\newcommand{\nquestions}{3284\xspace}
\newcommand{\nlmmsevaluated}{20\xspace}
\newcommand{\ntasks}{five\xspace}
\newcommand{\ngraphproperties}{23\xspace}
\newcommand{\highscore}{21.0}

\begin{document}
\maketitle

\begin{abstract}

Large multimodal models (LMMs) have exhibited proficiencies across many visual tasks. Although numerous well-known benchmarks exist to evaluate model performance, they increasingly have insufficient headroom. As such, there is a pressing need for a new generation of benchmarks challenging enough for the next generation of LMMs. One area that LMMs show potential is graph analysis, specifically, the tasks an analyst might typically perform when interpreting figures such as estimating the mean, intercepts or correlations of functions and data series. In this work, we introduce \datasetName, a graph analysis benchmark, fit for current and future frontier LMMs. Our benchmark is predominantly synthetic, ensuring high-quality, noise-free questions. \datasetName is comprised of \nquestions questions, covering \ntasks tasks and \ngraphproperties graph properties. We evaluate \nlmmsevaluated LMMs on \datasetName, finding it to be a challenging benchmark, with the highest performing model attaining a score of just \highscore\%. 
Finally, we conduct various ablations to investigate where the models succeed and struggle. We release \datasetName and a lightweight \datasetName-Lite to encourage progress in this important, growing domain.

\end{abstract}

\section{Introduction}

Driven by the union 
of increased compute resources, a rapidly growing research community, and competition fuelled by burgeoning %
commercial opportunities, the capabilities of frontier large multimodal models are swiftly improving and expanding. Furthermore, the timelines for advances, especially among closed-source models, appear to be shortening, with the time between model releases diminishing. This has important implications for evaluation and benchmarking, which are critical to gauge the relative strengths and weaknesses of models. In particular, as performance increases, common benchmarks become saturated and stale -- for example, GPT-4o \cite{gpt4o} scores 90.5\%, 90.2\%, and 88.7\% on the widely used MGSM, HumanEval, MMLU benchmarks, respectively. The usefulness of a benchmark in distinguishing model capability diminishes when there is little room for improvement. Moreover, label inaccuracies are reportedly pervasive \cite{northcutt2021pervasive} even in the most used benchmarks, further reducing possible headroom.

\begin{figure}[t]
\centering
\includegraphics[width=0.45\textwidth]{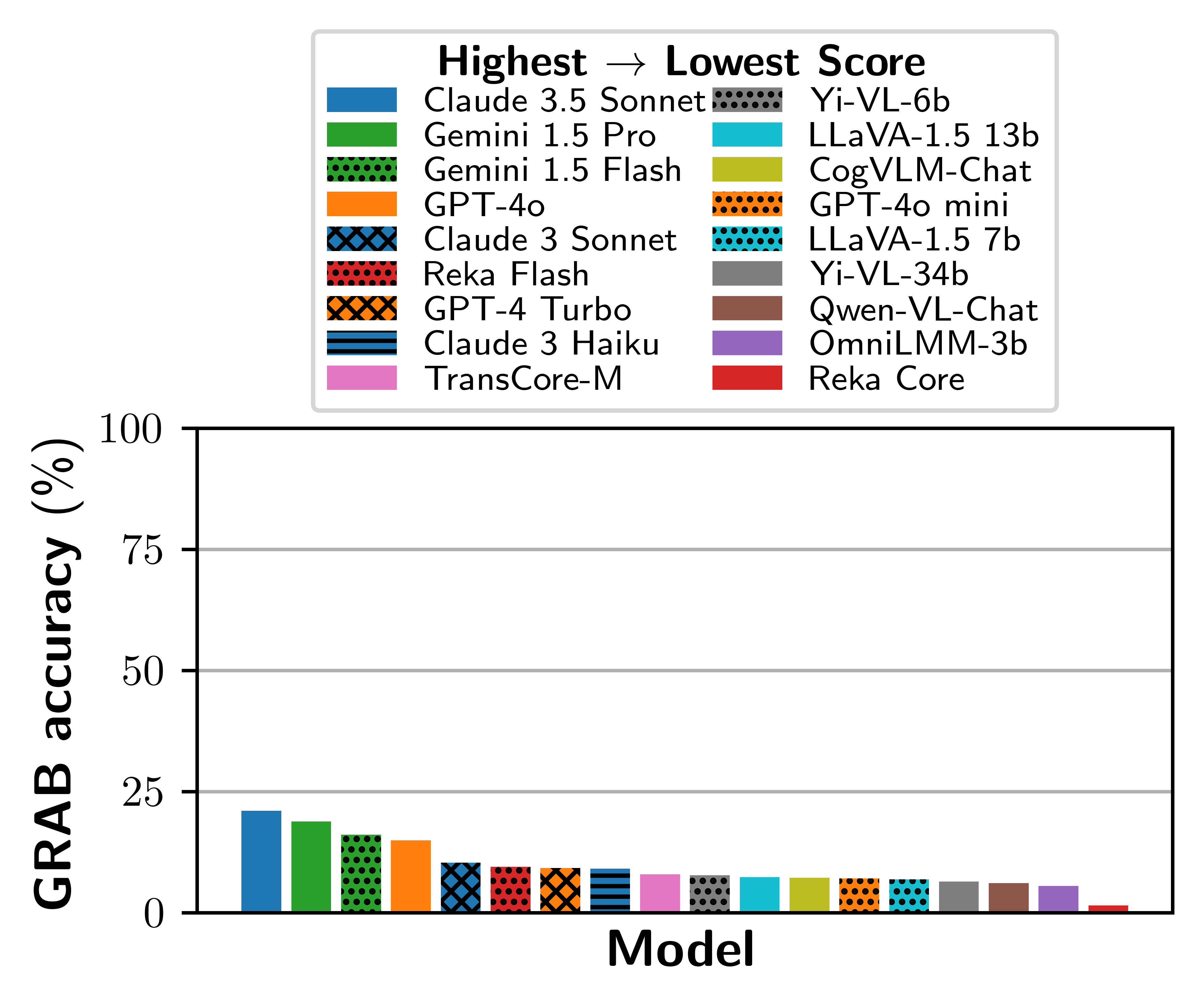}
\vspace{-0.4cm}
\caption{\textbf{Overall performance on GRAB}. Our benchmark proves challenging for frontier LMMs. The highest performing LMM attains an accuracy of just \highscore\% on GRAB.}
\label{fig:bar}
\end{figure}

\noindent\begin{figure*}[t]
\centering
    \includegraphics[width=\textwidth]{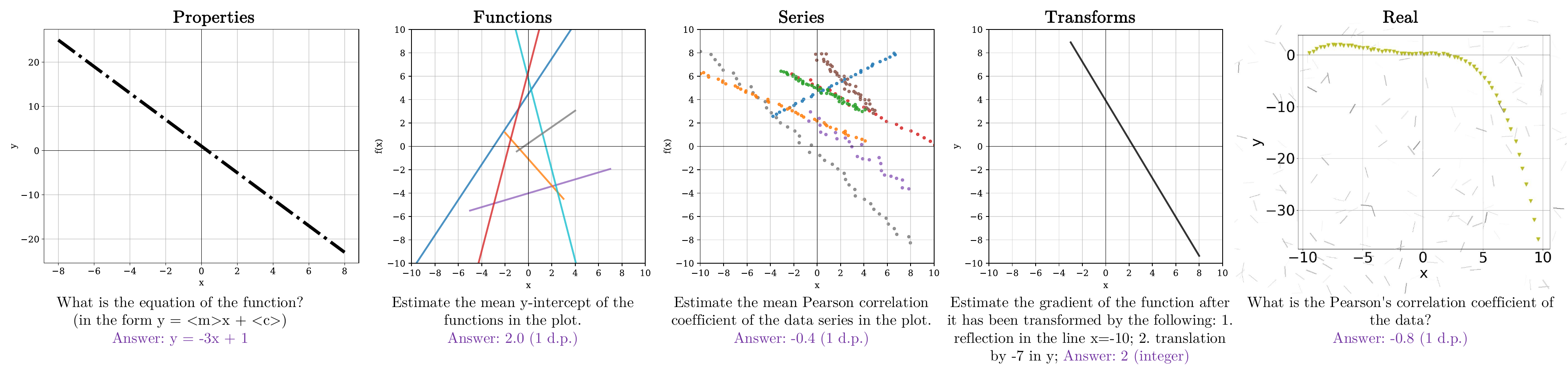}
\vspace{-0.7cm}
\caption{\textbf{The \underline{GR}aph \underline{A}nalysis \underline{B}enchmark}  consists of \nquestions graph analysis questions that prove challenging for frontier LMMs. The questions cover \ngraphproperties graph properties organised into \ntasks core tasks: (i)~\textbf{Properties} focuses on the analysis of features of individual functions and series; %
(ii)~\textbf{Functions} and (iii)~\textbf{Series} require computing the mean of properties across multiple functions and series; (iv)~\textbf{Transforms} involves determining the properties of a function after it has undergone a series of transforms; and, (v)~\textbf{Real} encompasses question styles from the other tasks in more realistic formats, including sketched on whiteboard or paper, embedded in digital contexts or with added noise.}%
\label{fig:overview}
\end{figure*}
One particular task of interest and importance is the interpretation of scientific and mathematical figures and charts, which is central to many analytical endeavours. Specifically, we target use cases where the underlying data of a figure is not accessible, for example in documents, sketches, or other image types, and quantities can only be deduced from visual interpretation. Given their prevalence, the insights gained from understanding and reasoning over these types of figures make this task an important and valuable capability for LMMs. Various forms of chart evaluation have been present in machine learning literature for quite some time through established benchmarks \cite{kahou2017figureqa,methani2020plotqa} and more recent works \cite{xu2023chartbench,roberts2024scifibench}. However, these benchmarks need an uplift for the current and future generations of frontier models.

To this end, we introduce \underline{GR}aph \underline{A}nalysis \underline{B}enchmark (GRAB), consisting of \nquestions questions. Almost all figures in GRAB are generated synthetically using the Matplotlib library \cite{plt}. Although some realism might be lost when focusing on synthetically generated data, there are numerous benefits to taking this approach: (1) \mbox{\textbf{Difficulty}} -- the complexity of questions can be controlled and tuned; (2) \mbox{\textbf{Noise}} -- ground truth is determined automatically during curation, avoiding the need for potentially erroneous post hoc annotation; (3) \mbox{\textbf{Composition}} -- we can directly select the characteristics of questions in our benchmark and arbitrarily control the scale; and, (4) \mbox{\textbf{Contamination}} -- while the same figure style will almost certainly have been seen during the pretraining of most LMMs, it is unlikely that the exact figures and questions generated for our benchmark have. The inclusion of a 1114-question `Real' task with figures that are hand-drawn, have added noise or are placed in digital screenshots increases the realism of the benchmark by introducing artefacts, distractors and camera effects.

Leveraging the control enabled by our synthetic generation, we optimise GRAB for usefulness and longevity. Specifically, we include challenging questions that frontier LMMs struggle to answer (see Fig.~\ref{fig:bar}) and we control the aesthetic parameters of each graph to ensure all relevant features are clear and questions are answerable. To balance streamlined evaluation and robustness, we include \nquestions questions in GRAB. We also provide a smaller 500-question version, called GRAB-Lite, as a lighter-weight evaluation.

We aim for a comprehensive evaluation of the abilities of frontier models to analyse figures by including a broad range of chart types and question formats. The tasks in GRAB involve determining key data properties, such as mean, variance, and interquartile range, as well as function properties like gradients, stationary points, and function parameters. Question complexity is increased by incorporating multiple functions and series (see Fig.~\ref{fig:overview}) and increasing the required answer precision.
We evaluate \nlmmsevaluated LMMs on our benchmark, which proves extremely challenging for the current generation of models, with the best performing attaining a score of \highscore\%. While some models exhibit a modicum of success on the easiest question set, the vast majority of our benchmark is beyond the capabilities of all models. We conduct a detailed error analysis and set of ablations to gain insights into which tasks and categories were the most challenging, as well as investigations of how question format and plotting library impact performance.

We summarise the major contributions of this work as follows:
(i) We introduce GRAB, %
a challenging \nquestions question benchmark for graph analysis, along with a smaller 500-question GRAB-Lite version 
(ii) We comprehensively characterise the capabilities of \nlmmsevaluated current generation frontier LMMs on GRAB, and (iii) We provide insights into model strengths and limitations through numerous ablations.

\section{Related Work}
\label{sec:related-work}

\subsection{Figure interpretation benchmarks}

Motivated by the last decade of progress in computer vision, a range of prior work has sought to evaluate figure interpretation capabilities of leading models. 
One family of work has focused on \textit{synthetic} data generation pipelines, profiting from the reliable ground truth and fine-grained control enabled by this approach.
This has yielded figure binary classification tasks~\cite{kahou2017figureqa}, captioning~\cite{chen2019figure}, and various structural understanding, retrieval and reasoning tasks with limited vocabularies~\cite{kafle2018dvqa}.
Synthetic figure generation has also been combined with corpora of online statistics~\cite{chaudhry2020leaf,singh2020stl,methani2020plotqa} and Kaggle datasets~\cite{xu2023chartbench} to achieve greater figure diversity.
A second line of work has sought to construct datasets by curating `real' figures directly from online sources and parsing their metadata.
For this purpose, figures have been sourced from research papers~\cite{siegel2016figureseer,roberts2024scifibench} and online sources such as \textit{Our World in Data}~\cite{masry2022chartqa}. \datasetName draws particular inspiration from the family of synthetic benchmarks, to which it also belongs. However, a key distinction between our tasks and prior works %
is GRAB's significantly greater difficulty, as evidenced by our empirical results illustrating that even the strongest frontier LMMs struggle to make headway (\S\ref{sec:experiments}).

\subsection{LMM evaluation and benchmarks}

Since the emergence of LMMs, numerous benchmarks have been created to evaluate and characterise their capabilities across many domains. Interpreting and answering questions of natural images is a key task, with benchmarks such as MMBench \cite{liu2023mmbench}, SEEDBench \cite{li2023seedbench}, SEEDBench-2 \cite{li2023seedbench2}, MME \cite{fu2023mme}, MM-Vet \cite{yu2023mm}, LVLM-eHub \cite{xu2023lvlm}, ScienceQA \cite{lu2022learn} and MMMU~\cite{yue2023mmmu} finding prominence. However, as outlined in~\cite{chen2024we}, although widely used, some of these most popular benchmarks contain problematic questions, including those that do not require visual content to be answered. Unlike older datasets developed to cater to the supervised learning paradigm that typically required much larger scales, more recent LMM-focused benchmarks contain significantly smaller test sets of a few thousand questions or a few hundred (\eg, \cite{padlewski2024vibe}). Various benchmarks can be found in more niche domains, such as geographic and geospatial evaluation \cite{roberts2023charting}, hallucinations \cite{guan2023hallusionbench} and OCR \cite{liu2023hidden}. A closely related work to GRAB is MathVista~\cite{lu2023mathvista}, which evaluates mathematical reasoning in visual contexts. MathVista is a metadataset (similar to \cite{roberts2023satin}) consisting of a broad range of mostly multiple-choice questions from 31 multimodal datasets covering a diverse set of figure types and mathematical tasks ranging from table interpretation to puzzles. Though similar, the focus of MathVista on general mathematical reasoning, is distinct from ours, which is centered on the more narrow task of deriving and analysing graph properties. Moreover, our benchmark is considerably more challenging: GPT-4o scores >60\% on MathVista\footnote{\url{https://huggingface.co/spaces/opencompass/open_vlm_leaderboard}} compared to 14.9\% on GRAB.

\section{GRaph Analysis Benchmark}

\newcommand{\linebrightness}{0.7}

\begin{figure*}[t]
\centering
\newlength{\targetheight}
\setlength{\targetheight}{1.9cm}

\begin{minipage}[t]{0.73\textwidth}
    \vspace{0pt} %
    \adjustbox{height=\targetheight}{%
        \begin{tabular}{ll}
        \hline
        \textbf{Statistic} & \textbf{Number} \\
        \hline
        Questions & \nquestions \\
        Tasks & 5 \\
        Graph properties & \ngraphproperties \\
        \arrayrulecolor[gray]{\linebrightness} \hline
        \textit{Properties} questions & 660 \\
        \textit{Functions} questions & 710 \\
        \textit{Series} questions & 490 \\
        \textit{Transforms} questions & 310 \\
        \textit{Real} questions & 1114 \\
        \arrayrulecolor[gray]{\linebrightness} \hline
        10s precision questions & 46\\
        Integer precision questions & 2450\\
        1 d.p. precision questions & 787\\
        \arrayrulecolor[gray]{0.1} \hline
        \end{tabular}%
    }%
    \hfill
    \adjustbox{height=\targetheight}{%
        \begin{tabular}{p{4cm}p{6cm}}
        \hline
        \textbf{Category} & \textbf{Properties} \\
        \hline
        Intercepts \& Gradients & x-intercept, y-intercept, gradient \\
        \arrayrulecolor[gray]{\linebrightness} \hline
        Stationary Points & stationary point(s) coordinates \\
        \arrayrulecolor[gray]{\linebrightness} \hline
        Trigonometric & amplitude, vertical shift, period \\
        \arrayrulecolor[gray]{\linebrightness} \hline
        Functions & function equations \\
        \arrayrulecolor[gray]{\linebrightness} \hline
        Counting & number of points, number of series \\
        \arrayrulecolor[gray]{\linebrightness} \hline
        Correlation & \{Pearson's, Spearman's rank, Kendall rank\} correlation coefficient \\
        \arrayrulecolor[gray]{\linebrightness} \hline
        Area Bounded & total area bounded, net area bounded \\
        \arrayrulecolor[gray]{\linebrightness} \hline
        Measures of Spread & mean, median, interquartile range, variance \\
        \arrayrulecolor[gray]{\linebrightness} \hline
        Range \& Extrema & min/max values, domain length, range \\
        \arrayrulecolor[gray]{0.1} \hline
        \end{tabular}%
    }
    
    \vspace{0.5em}
    
    \begin{minipage}{0.35\textwidth}
        \centering
        \vspace{-0.05cm}
        \captionof{table}{\textbf{GRAB statistics}}
        \label{tab:grab_statistics}
    \end{minipage}%
    \hfill
    \begin{minipage}{0.6\textwidth}
        \centering
        \vspace{-0.05cm}
        \captionof{table}{\textbf{GRAB categories and graph properties}}
        \label{tab:categories_and_properties}
    \end{minipage}
\end{minipage}%
\hfill
\begin{minipage}[t]{0.26\textwidth}
    \vspace{0.1cm} %
    \raggedleft
    \adjustbox{height=3.7cm,valign=t}{%
        \includegraphics{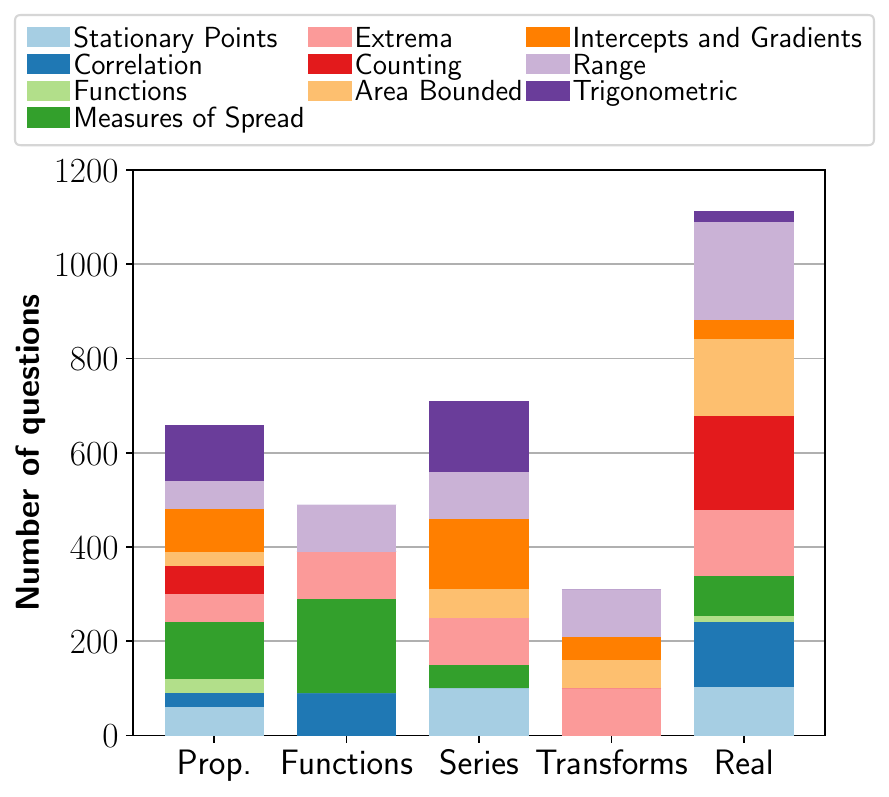}%
    }
    \vspace{-0.3cm}
    \caption{\textbf{GRAB categories}}
    \label{fig:cat}
\end{minipage}
\end{figure*}

Our GRaph Analysis Benchmark consists of \nquestions questions covering \ntasks core tasks and \ngraphproperties graph properties. In the following section, we provide an overview of the benchmark, including descriptions of the categories and tasks, example questions and answers, and an overview of our synthetic curation process. High-level statistics can be found in Tab.~\ref{tab:grab_statistics}.

\subsection{Categories}

Questions in our GRAB benchmark cover 23 graph properties from which analytical insights can commonly be derived from data or functions in graphs and charts. We group these properties into the categories shown in Tab.~\ref{tab:categories_and_properties}. We consider this set of properties to cover most real use cases and require some measure of visual analytical reasoning. We refrain from including `easier' OCR-centered questions -- such as legend, title, axis-label interpretation -- in our benchmark to keep the focus on visual analytical reasoning, and instead only include questions of this type as an ablation. Furthermore, as outlined in \S\ref{sec:related-work}, numerous prior benchmarks have included this type of question.

\subsection{Tasks}

\begin{figure*}
\centering
        \includegraphics[width=0.9\textwidth]{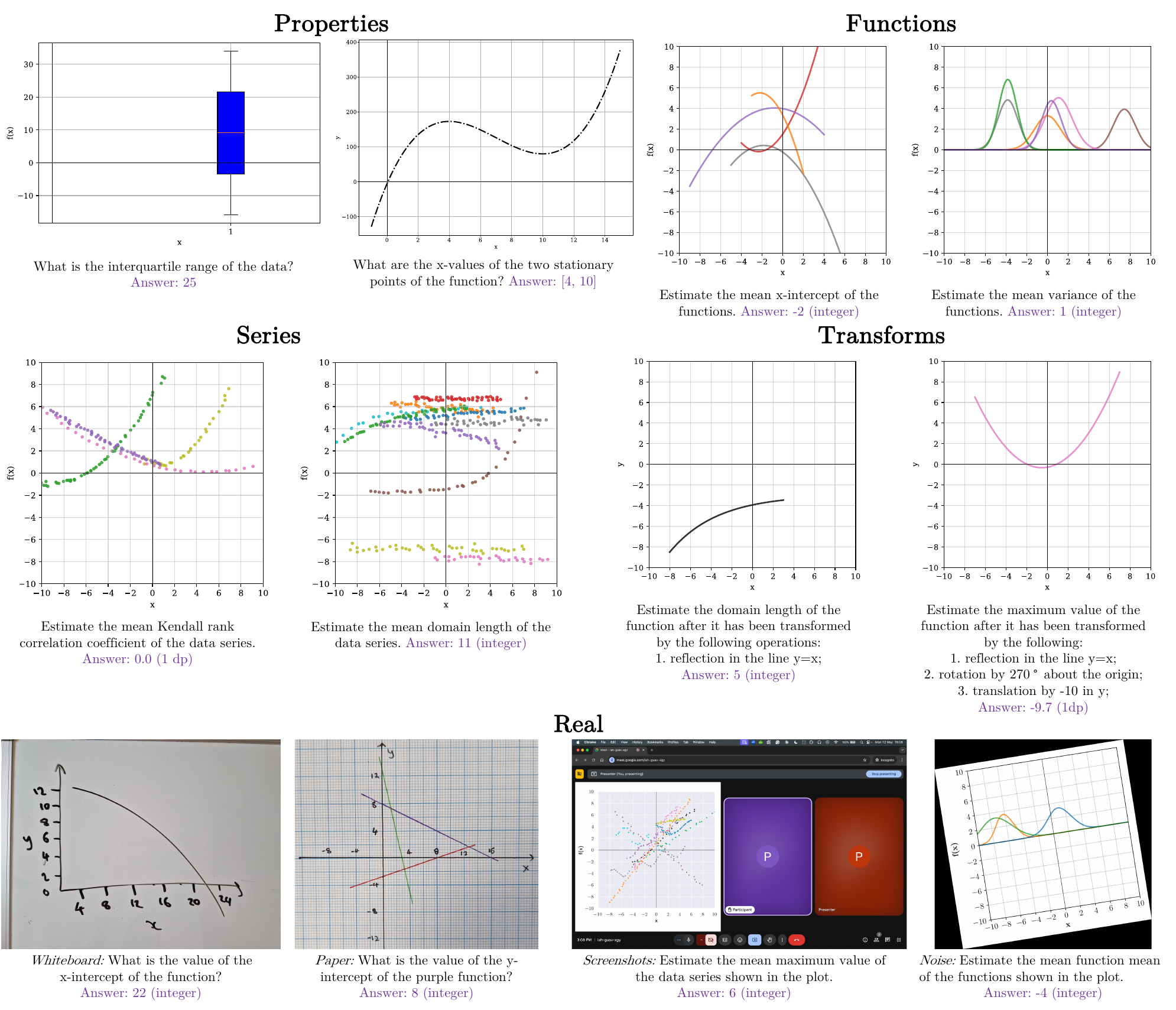}
    \vspace{-0.3cm}
    \caption{\textbf{Example questions for the five tasks in the GRAB benchmark.} All questions include synthetically rendered graphs apart from the Whiteboard and Paper splits of the \textit{Real} task, which are hand-drawn and photographed.}
    \label{fig:examples0}
\end{figure*}

We structure GRAB into the following tasks covering different aspects of graph analysis. Example questions and answers for each task can be found in Figs.~\ref{fig:overview} and ~\ref{fig:examples0}.

\subsubsection{\textit{Properties}}
This initial task aims to provide a broad evaluation of the capability to understand and derive properties of graphs, including both basic fundamental properties like \textit{number of series} or more advanced properties requiring calculations, like \textit{total area bounded}. Aside from \textit{number of series} questions, the graphs involve just a single function or data series.

\subsubsection{\textit{Functions}}
\textit{Functions} builds on \textit{Properties} with increased complexity by adding up to a total of 10 functions per graph. Questions %
require an additional step of averaging across all functions. For example, `Estimate the mean total area bounded by the function and the x-axis of the functions shown in the plot'. Added difficulty is introduced by functions overlapping.

\subsubsection{\textit{Series}}

Similar to \textit{Functions}, \textit{Series} builds on the initial task by adding up to 10 data series per graph, with questions requiring estimates of the mean of specific properties across the observable series, for example, `Estimate the mean Pearson correlation coefficient of the data series shown in the plot'. The additional series introduces challenges such as overlapping and clustered points. 

\subsubsection{\textit{Transforms}}

Questions in the \textit{Transforms} task include a single function and require computing a graph property after performing a sequence of up to 10 transforms (rotations, translations, scales and reflections).

\subsubsection{\textit{Real}}

Organised into four splits, the \textit{Real} task modifies the other tasks with increased realism. This task includes \textit{Properties}-style questions with figures hand-drawn on (1) \textbf{Paper} and (2) \textbf{Whiteboards}, and questions styled on all the other tasks with figures placed in (3) \textbf{Screenshots} (\textit{e.g.}, in emails, slides, documents and video calls) or with arbitrary (4) \textbf{Noise} added to the figures (\textit{e.g.}, artefacts, flips and blurs).

\subsection{Curation}

We leverage the Matplotlib \cite{plt} and Seaborn \cite{Waskom2021} libraries to synthetically create the majority of the graphs in GRAB and hand-draw only a small subset of the \textit{Real} task. %
For each task, we initially created a longlist of 250 questions per graph property. We then performed downsampling to select a final question set with uniformly distributed answers. This downselection aimed to increase question diversity, reduce bias in the generation process and avoid an overabundance of questions with answers converging towards 0. %
We associate each graph property (Tab.~\ref{tab:categories_and_properties}) with a set of data generation functions and function parameter ranges. For categories such as \textit{Intercepts \& Gradients} or \textit{Functions}, we sample values for the property of interest and then fit the generation parameters to produce functions/series that match the value. For the other categories, such as \textit{Correlation} or \textit{Area Bounded}, we randomly sample the generation parameters and calculate the property value from the data. In each case, we ensure questions are unique. For each graph in the \textit{Properties} task, aesthetic parameters (\eg figure/font size and line style) are randomly sampled from a defined set of options. For the \textit{Series}, \textit{Functions} and \textit{Transforms} tasks, we limit the variability of the graphs by configuring the appearance to a defined schema: all graphs have the same x and y-limits, grid lines, tick labels, font and resolution (2251x2171). A single function type is used for each question, irrespective of the number of series. Synthetically generated questions from the \textit{Real} task leveraged the generation procedure from the other tasks to create new figures. These figures were then embedded in digital screenshots (\textbf{Screenshots} split) or manipulated with visual noise (\textbf{Noise} split). Figures in the \textbf{Paper} and \textbf{Whiteboard} splits were hand-drawn using a range of functions and graph styles and subsequently photographed. 

For each task, we include sufficient questions to represent different combinations of graph properties and question complexity. We aim for an approximate balance per graph property, resulting in the category distribution shown in Fig. \ref{fig:cat}. Most questions require answers to integer precision, though some categories (\textit{e.g.,} \textit{Area Bounded}) are evaluated to nearest 10 precision. To increase the difficulty, $\sim$25\% of questions require 1 decimal place precision.

\subsubsection{Quality control}

The curation process underwent numerous iterations of human review that sought to tune the hyperparameters involved in creating the graphs. This activity was primarily focused on jointly ensuring (i) the questions were answerable from a technical perspective, (ii) the ground-truth answers were correct and that (iii) the rendered figures were suitably readable. Similar spot-checks were carried out on the final dataset. %

\section{Experiments}
\label{sec:experiments}

\subsection{Baselines}

We evaluate \nlmmsevaluated LMMs on GRAB, covering a broad range of both open- and closed-source models. We use chat or instruction-tuned variants rather than base models, of each model evaluated, as they are typically better at following task and output instructions. We benchmark 4 families of frontier closed-source LMMs, specifically the GPT-4 \cite{gpt4v,gpt4o}, Gemini \cite{team2023gemini, reid2024gemini} Claude 3 \cite{claude3}, and Reka \cite{ormazabal2024reka} series. Additionally, we select the following open-source models for evaluation: Qwen-VL \cite{bai2023qwen}, TransCore-M \cite{TransCore_M}, OmniLMM \cite{omnilmm}, Yi \cite{yi}, CogVLM \cite{wang2023cogvlm}, and LLaVA-1.5 \cite{liu2024improved}. %

\subsection{Experimental settings}

\subsubsection{Inference}
We evaluate the closed-source models via the Vertex AI API \cite{vertexaiapi} \{Gemini, Claude\}, OpenAI API \cite{openaiapi} \{GPT\}, and Reka API \cite{rekaapi} \{Reka\}. Following benchmarking conventions, we set model hyperparameters that result in deterministic and reproducible generation. This entails utilising a greedy search decoding strategy in which the most probable token is selected from the model vocabulary $V$ at each step, conditional on the preceding tokens \ie, $w_{n+1} = \arg\max_{w \in V} P(w | w_1, w_2, \ldots, w_n)$. We achieve this by setting the $temperature$ and $top k$ parameters to zero and 1, respectively, and specifying random seeds. We run the open-source models using HuggingFace Transformers \cite{wolf-etal-2020-transformers} and the OpenCompass toolkit \cite{2023opencompass}.%

\subsubsection{Prompting}

We adopt a simple prompting strategy throughout our evaluation, similar to that used by the popular repository \textit{simple-evals} \cite{simpleevals}. Concretely, we use a single basic \textit{user} prompt consisting of the question and output format instructions. We do not modify the system prompt or tailor the prompt for each model. %
Although model performance is known to be sensitive to these input configurations, our investigation focuses on chat or instruction-tuned models, which are optimised for instruction following: \textit{if a model can only perform a task when a specific phrase is included in the prompt, this is a clear weakness.} Being able to answer questions in a realistic setting (\ie as a human would be presented with the questions) offers a more generalised evaluation. %

\subsubsection{Evaluation}
\label{sec:evaluation}
Our core evaluation protocol is %
exact matching between model output and the ground truth answers. The only post-processing step we take is removing leading whitespace. An answer is only scored as correct if the output characters at each position match those of the expected answer. This means that although verbose answers of the form `\textit{The answer is...}' %
or `<Reasoning> <Answer>' might contain the correct answer, they will be marked incorrect. While this may lower model scores compared to human or LLM evaluation (\eg \cite{roberts2024scifibench, padlewski2024vibe}), we consider this a better evaluation as it jointly assesses task proficiency and instruction-following ability. Furthermore, this methodology does not necessitate (potentially) timely human scoring or the means to run a strong LLM evaluator. As mentioned previously, when considering potential downstream usage, a model is not useful if it can perform a given reasoning task perfectly but fails to reliably follow output instructions. As an ablation, we compare this strict exact matching protocol to automatic evaluation using both closed-source and open-source LLMs.

\subsection{Overall results}

\begin{table}[t]
\resizebox{1\columnwidth}{!}{
\centering
\begin{tabular}{lllllll}
\hline
                & \multicolumn{6}{c}{\textbf{Accuracy (\%)}}               \\
                & \textit{Properties} & \textit{Functions} & \textit{Series} & \textit{Transforms} & \textit{Real} & \textbf{Overall}\\
\textbf{Model}  &        (660)        &           (710)         & (490)          & (310)    & (1114)   & (3284) \\
\hline
\multicolumn{7}{c}{\light{Closed-source LMMs}} \\
Claude 3 Haiku   \cite{claude3}               & 14.2 & 6.6  & 8.8  & 3.9  & 9.2  & 9.1  \\
Claude 3 Sonnet \cite{claude3}                & 15.3 & 8.6  & 4.5  & 4.8  & 12.4 & 10.3 \\
Claude 3.5 Sonnet \cite{claude3.5}            & \cellcolor{table_shade}\textbf{41.8} & \cellcolor{table_shade}\textbf{15.5} & 11.0 & \cellcolor{table_shade}\textbf{10.0} & 19.6 & \cellcolor{table_shade}\textbf{21.0} \\
GPT-4 Turbo \cite{gpt4v}                      & 18.5 & 8.5  & 4.9  & 3.5  & 7.5  & 9.2  \\
GPT-4o mini \cite{gpt4o_mini}                 & 15.8 & 6.8  & 5.7  & 2.9  & 4.0  & 7.1  \\
GPT-4o \cite{gpt4o}                           & 24.7 & 10.8 & 9.2  & 3.5  & 17.3 & 14.9 \\
Gemini 1.0 Pro Vision   \cite{team2023gemini} & 20.2 & 5.8  & 6.9  & 6.1  & $^*$    & $^*$    \\
Gemini 1.5 Flash \cite{reid2024gemini}        & 28.5 & 11.5 & 8.4  & 9.0  & 17.1 & 16.1 \\
Gemini 1.5 Pro \cite{reid2024gemini}          & 34.2 & 11.4 & \cellcolor{table_shade}\textbf{13.3} & 6.5  & \cellcolor{table_shade}\textbf{20.3} & 18.8 \\
Reka Edge \cite{ormazabal2024reka}            & 11.8 & 8.7  & 11.6 & 1.9  & $^*$    & $^*$    \\
Reka Flash \cite{ormazabal2024reka}           & 13.2 & 10.1 & 6.3  & 3.9  & 10.0 & 9.5  \\
Reka Core \cite{ormazabal2024reka}            & 1.7  & 0.0  & 4.3  & 0.3  & 1.3  & 1.5  \\
\arrayrulecolor[gray]{\linebrightness} \hline
\multicolumn{7}{c}{\light{Open-source LMMs}} \\
CogVLM-Chat \cite{wang2023cogvlm}             & 7.0  & 4.9  & 5.1  & 3.9  & \cellcolor{table_shade}\textbf{10.5} & 7.2  \\
Qwen-VL-Chat \cite{bai2023qwen}               & \cellcolor{table_shade}\textbf{10.2} & 6.6  & 5.1  & 2.9  & 4.6  & 6.1  \\
OmniLMM-3b \cite{omnilmm}                     & 6.7  & 4.9  & 4.1  & 4.5  & 6.2  & 5.5  \\
TransCore-M \cite{TransCore_M}                & 7.9  & \cellcolor{table_shade}\textbf{9.2}  & 7.6  & 3.9  & 8.2  & \cellcolor{table_shade}\textbf{7.9}  \\
Yi-VL-6b \cite{yi}                            & 5.6  & 8.6  & 7.1  & 4.2  & 9.7  & 7.7  \\
Yi-VL-34b \cite{yi}                           & 7.6  & 5.9  & 5.5  & 2.3  & 7.5  & 6.4  \\
LLaVA-1.5 7b \cite{liu2024improved}           & 4.7  & 7.5  & 6.5  & \cellcolor{table_shade}\textbf{4.8} & 8.5  & 6.9  \\
LLaVA-1.5 13b   \cite{liu2024improved}        & 5.0  & 7.7  & \cellcolor{table_shade}\textbf{8.4} & 3.9  & 8.9  & 7.3  \\
\arrayrulecolor[gray]{0.1} \hline
\end{tabular}}
\vspace{-0.2cm}
\caption{\textbf{Results on GRAB}. The highest open and closed-source model scores for each task are \colorbox{table_shade}{highlighted} and \textbf{bold}. $^*$ unable to perform full evaluation due to model deprecation. Overall scores are weighted averages based on task size.}
\label{tab:main_results}
\end{table}

We present the results on GRAB in Tab.~\ref{tab:main_results}. A clear observation is that \textbf{our benchmark proves challenging for frontier models}, with low overall performances from all models (see also Fig. \ref{fig:bar}). \textbf{The highest performing model, Claude 3.5 Sonnet, achieves a score of just \highscore\%}.
As expected, the highest scores are attained on the \textit{Properties} task, which includes questions that just require the derivation of a single graph property. Lower scores are achieved on the \textit{Functions} and \textit{Series} tasks, likely due to the added complexity of computing the mean of a given property over multiple functions/series. There is a macro-trend of slightly higher accuracy on the functions task compared to the series task, possibly due to the irregularity introduced by the noise in the series data points. Significantly lower accuracy is attained on the \textit{Transforms} task. Reasonable scores are attained on the \textit{Real} task, which includes questions from the other tasks, that approximately correspond to the average scores on the other tasks. This indicates that the \textit{Real} questions are of a similar difficulty. 

A contributing factor to these scores is the degree to which the models follow output format instructions. Given our exact string matching evaluation protocol, any deviation from the expected output is deemed incorrect. Despite potentially having superior mathematical reasoning abilities, models that struggle to precisely follow instructions or have proclivities to generate verbose answers suffer severe performance degradation. With the exception of Reka Core (which showed poor output format instruction following capabilities), all closed-source models outperformed the leading open-source model. However, the open-source models did consistently generate answers in the correct format. In this low-scoring regime where even the best models struggle, a comparison of model performance should be made with the caveat that all models can barely perform the tasks in the benchmark. \textbf{We anticipate our benchmark will yield most benefit when evaluating the \textit{next} generation of LMMs}.

\subsection{Real splits analysis}

In Tab. \ref{tab:grab_real}, we report model scores on questions in each split in the \textit{Real} task. The highest scores are attained on the Whiteboard and Paper splits, which include only less complex \textit{Properties}-style questions. Notably, Claude 3.5 Sonnet struggles, scoring higher on the other splits. In general, model scores on both the Screenshots and Noise splits are comparable to the overall GRAB scores. This suggests the models are well-suited to dealing with real-world images, as the addition of visual distortions due to the noise artefacts or distractors and additional context in the screenshots does not significantly degrade performance. We also report the performance of two newer releases of the Gemini 1.5 Flash model (2.0 and 2.5), each showing subsequent improvements.

\begin{table}[t]
\resizebox{1\columnwidth}{!}{
\centering
\begin{tabular}{llllll}
\hline
                & \multicolumn{5}{c}{\textbf{Accuracy (\%)}}               \\
                & \textit{Whiteboard} & \textit{Paper} & \textit{Screenshots} & \textit{Noise} & \textbf{Overall}\\
\textbf{Model}  &        (41)        &           (41)         & (516)          & (516)       & (1114) \\
\hline
\multicolumn{6}{c}{\light{Closed-source LMMs}} \\
Claude 3 Haiku   \cite{claude3}               & 9.8 & 4.9 & 9.5 & 9.3 & 9.2 \\
Claude 3 Sonnet \cite{claude3}                & 9.8 & 14.6 & 12.8 & 12.0 & 12.4 \\
Claude 3.5 Sonnet \cite{claude3.5} & 14.6 & 17.0 & 18.6 & \cellcolor{table_shade}\textbf{21.1} & 19.6 \\
GPT-4 Turbo \cite{gpt4v}                      & 9.8 & 4.9 & 6.6 & 8.3 & 7.5 \\
GPT-4o mini \cite{gpt4o_mini}                 & 7.3 & 2.4 & 4.7 & 3.3 & 4.0 \\
GPT-4o \cite{gpt4o}                           & 29.3 & 9.8 & 18.0 & 16.3 & 17.3 \\
Gemini 1.5 Flash \cite{reid2024gemini}        & 31.7 & 17.1 & 16.3 & 16.7 & 17.1 \\
Gemini 1.5 Pro \cite{reid2024gemini}          & \cellcolor{table_shade}\textbf{34.4} & \cellcolor{table_shade}\textbf{24.4} & \cellcolor{table_shade}\textbf{21.9} & 17.2 & \cellcolor{table_shade}\textbf{20.3} \\
Reka Flash \cite{ormazabal2024reka}           & 19.5 & 7.3 & 7.9 & 11.4 & 10.0 \\
Reka Core \cite{ormazabal2024reka}            & 4.9 & 0.0 & 1.0 & 1.4 & 1.3 \\
\arrayrulecolor[gray]{\linebrightness} \hline
\multicolumn{6}{c}{\light{Open-source LMMs}} \\
CogVLM-Chat \cite{wang2023cogvlm}             & \cellcolor{table_shade}\textbf{14.6} & 7.3 & \cellcolor{table_shade}\textbf{11.0} & 9.9 & \cellcolor{table_shade}\textbf{10.5} \\
Qwen-VL-Chat \cite{bai2023qwen}               & 12.2 & 4.9 & 2.9 & 5.6 & 4.6 \\
OmniLMM-3b \cite{omnilmm}                     & 7.3 & 2.4 & 6.6 & 6.0 & 6.2 \\
TransCore-M \cite{TransCore_M}                & 2.4 & \cellcolor{table_shade}\textbf{9.8} & 7.6 & 9.1 & 8.2 \\
Yi-VL-6b \cite{yi}                            & 7.3 & 4.9 & 9.9 & \cellcolor{table_shade}\textbf{10.1} & 9.7 \\
Yi-VL-34b \cite{yi}                           & 4.9 & 4.9 & 7.8 & 7.8 & 7.5 \\
LLaVA-1.5 7b \cite{liu2024improved}           & 4.9 & 4.9 & 9.5 & 8.1 & 8.5 \\
LLaVA-1.5 13b   \cite{liu2024improved}        & 7.3 & 4.9 & 8.9 & 9.3 & 8.9 \\
\hline
\hline
Gemini 2.0 Flash \cite{gemini20flash} & 22.0 & 17.0 & 18.8 & 20.5 & 19.6 \\
Gemini 2.5 Flash \cite{gemini25flash} & \cellcolor{table_shade}\textbf{36.6} & \cellcolor{table_shade}\textbf{22.0} & \cellcolor{table_shade}\textbf{29.5} & \cellcolor{table_shade}\textbf{29.3} & \cellcolor{table_shade}\textbf{29.4} \\
\arrayrulecolor[gray]{0.1} \hline
\end{tabular}}
\vspace{-0.2cm}
\caption{\textbf{Performance on the GRAB `Real' task splits.} Overall scores are weighted averages based on task size.}
\label{tab:grab_real}
\end{table}

\subsection{Category analysis}

\def\catcolwidth{1.7cm}
\def\cathalfcolwidth{1cm}

\begin{table*}[t]
\centering
\resizebox{0.96\textwidth}{!}{
\begin{tabular}
{lp{\cathalfcolwidth}p{\catcolwidth}p{\cathalfcolwidth}p{\catcolwidth}p{\catcolwidth}p{\catcolwidth}p{\catcolwidth}p{\cathalfcolwidth}p{\cathalfcolwidth}p{1.5cm}}
\hline
      & \multicolumn{10}{c}{\textbf{Accuracy (\%)}}               \\
 \multicolumn{1}{r}{\textbf{Category}} & \textit{Range} & \textit{Measures of Spread} & \textit{Area Bounded} & \textit{Intercepts \& Gradients} & \textit{Stationary Points} & \textit{Functions} & \textit{Correlation} & \textit{Extrema} & \textit{Counting} & \textit{Trigonometric} \\
\textbf{Model} & (360)              & (370) & (150)                    & (290)      & (160)       & (30)             & (120)     & (360)   & (60)        & (270)         \\
\hline
\multicolumn{11}{c}{\light{Closed-source LMMs}} \\
Claude 3 Haiku   \cite{claude3}               & 4.7  & 11.9 & 2.7 & 11.4 & 10.6 & 0.0 & 20.8 & 5.8  & 13.3 & 10.0 \\
Claude 3 Sonnet \cite{claude3}                & 4.7  & 10.0 & 4.0 & 13.8 & 8.1  & 0.0 & 9.2  & 8.9  & 18.3 & 11.9 \\
Claude 3.5 Sonnet \cite{claude3.5}            & \cellcolor{table_shade}\textbf{13.6} & \cellcolor{table_shade}\textbf{28.6} & 2.7 & \cellcolor{table_shade}\textbf{25.5} & \cellcolor{table_shade}\textbf{25.6} & 0.0 & 9.2  & \cellcolor{table_shade}\textbf{24.7} & 30.0 & \cellcolor{table_shade}\textbf{29.3} \\
GPT-4 Turbo \cite{gpt4v}                      & 6.7  & 13.0 & 0.0 & 13.1 & 11.9 & 0.0 & 5.0  & 8.6  & 18.3 & 14.8 \\
GPT-4o mini \cite{gpt4o_mini}                 & 3.3  & 10.8 & 2.0 & 10.7 & 8.1  & 0.0 & 2.5  & 8.1  & 26.7 & 15.6 \\
GPT-4o \cite{gpt4o}                           & 9.2  & 18.6 & 2.0 & 14.1 & 16.2 & 0.0 & 15.8 & 8.1  & 30.0 & 21.5 \\
Gemini 1.0 Pro Vision   \cite{team2023gemini} & 4.7  & 16.2 & 1.3 & 7.9  & 13.1 & 0.0 & 12.5 & 13.1 & 23.3 & 10.4 \\
Gemini 1.5 Flash \cite{reid2024gemini}        & 11.7 & 18.9 & 2.7 & 16.2 & 16.9 & 0.0 & 8.3  & 22.2 & 25.0 & 16.3 \\
Gemini 1.5 Pro \cite{reid2024gemini}          & 9.4  & 23.2 & \cellcolor{table_shade}\textbf{4.7} & 17.9 & 24.4 & 0.0 & 26.7 & 19.4 & \cellcolor{table_shade}\textbf{33.3} & 19.3 \\
Reka Edge \cite{ormazabal2024reka}            & 2.8  & 9.7  & 3.3 & 10.0 & 8.1  & 0.0 & \cellcolor{table_shade}\textbf{38.3} & 4.7  & 18.3 & 13.3 \\
Reka Flash \cite{ormazabal2024reka}           & 6.7  & 11.6 & 3.3 & 15.5 & 7.5  & 0.0 & 7.5  & 5.3  & 18.3 & 12.6 \\
Reka Core \cite{ormazabal2024reka}            & 0.0  & 0.0  & 0.0 & 0.0  & 1.2  & 0.0 & 25.0 & 0.3  & 0.0  & 0.0  \\
\arrayrulecolor[gray]{\linebrightness} \hline
\multicolumn{11}{c}{\light{Open-source LMMs}} \\
CogVLM-Chat \cite{wang2023cogvlm}             & 6.9  & 5.4  & 2.7 & 4.1  & 5.6  & 0.0 & 0.8  & 4.7  & 16.7 & 7.4  \\
Qwen-VL-Chat \cite{bai2023qwen}               & 3.6  & \cellcolor{table_shade}\textbf{8.6}  & 1.3 & 6.6  & \cellcolor{table_shade}\textbf{10.6} & 0.0 & 1.7  & \cellcolor{table_shade}\textbf{6.7}  & \cellcolor{table_shade}\textbf{18.3} & 10.4 \\
OmniLMM-3b \cite{omnilmm}                     & 8.9  & 3.2  & 2.7 & 5.9  & 3.1  & 0.0 & 0.0  & 5.0  & 11.7 & 6.7  \\
TransCore-M \cite{TransCore_M}                & 8.6  & 4.1  & 2.7 & \cellcolor{table_shade}\textbf{10.0} & 8.8  & 0.0 & 15.8 & 5.8  & 8.3  & 10.4 \\
Yi-VL-6b \cite{yi}                            & 5.0  & 6.5  & \cellcolor{table_shade}\textbf{3.3} & 7.9  & 5.6  & 0.0 & 12.5 & 5.6  & 1.7  & \cellcolor{table_shade}\textbf{11.5} \\
Yi-VL-34b \cite{yi}                           & 2.8  & 7.8  & 2.7 & 8.3  & 1.9  & 0.0 & 8.3  & 4.2  & 6.7  & 10.0 \\
LLaVA-1.5 7b \cite{liu2024improved}         & \cellcolor{table_shade}\textbf{9.7} & 4.9 & \cellcolor{table_shade}\textbf{3.3} & 4.8 & 5.6 & 0.0 & 9.2  & 5.3 & 5.0  & 6.3 \\
LLaVA-1.5 13b   \cite{liu2024improved}        & 8.3 & 4.6 & 2.7 & 7.9 & 4.4 & 0.0 & \cellcolor{table_shade}\textbf{18.3} & 5.3 & 8.3 & 5.2 \\
\arrayrulecolor[gray]{0.1} \hline
\end{tabular}}
\vspace{-0.3cm}
\caption{\textbf{GRAB per-category performance.} The highest open and closed-source model scores for each category are \colorbox{table_shade}{highlighted} and \textbf{bold}. To isolate per-category performance and avoid the effects of noise and distractors, the results in this table do not include the \textit{Real} task.} %
\label{tab:category_results}
\end{table*}

A decomposition of model performance into question category is provided in Tab.~\ref{tab:category_results}. Significant variance is shown across categories, with some models much better suited for certain categories than others. Overall, the highest accuracies are attained on the \textit{Counting} category, which encompasses simple questions determining the number of series shown in each plot, or the approximate number of data points. On the other hand, \textbf{no model correctly answered a single question from the \textit{Functions} category}. This challenging category requires correct derivation of multiple parameters (\eg the gradient and intercept for linear functions) \textit{and} correct formatting of the output. Inspection of the model output suggests adherence to the strict output format was not an issue: the models were unable to correctly determine the function parameters. However, in a number of cases, proposed functions were close to the correct answers, \eg, with all parameters correct apart from one that was off by one. Minor perturbations to the graph appearance did result in a small number of correct answers.

\subsection{Complexity analysis}

\begin{figure*}[t]
\centering
\includegraphics[width=1.0\textwidth]{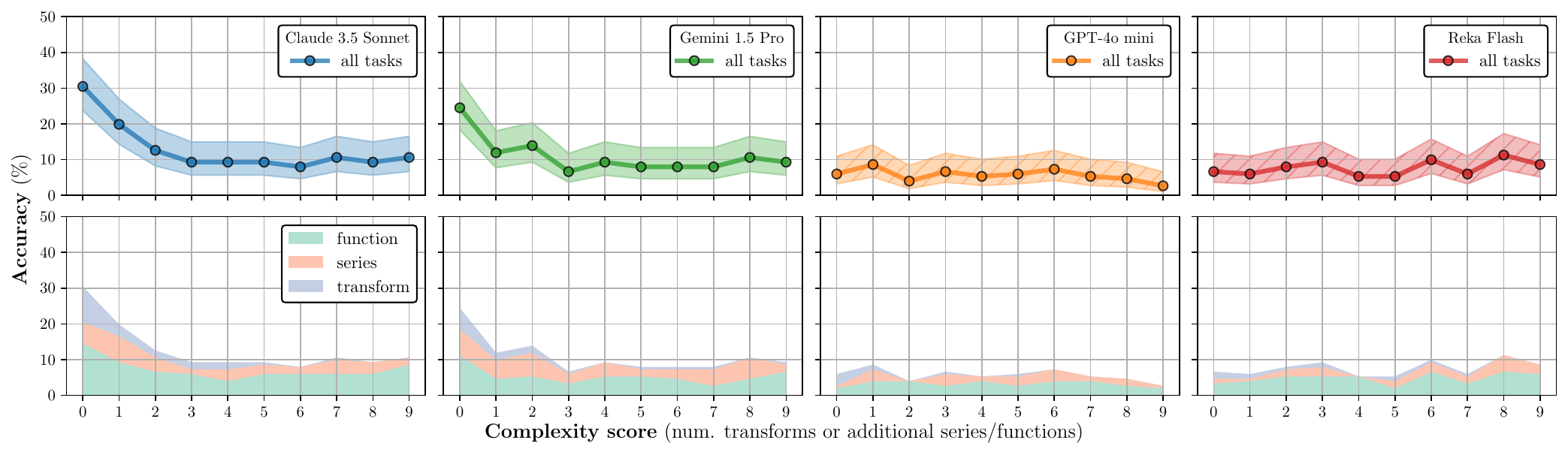}
\vspace{-0.5cm}
\caption{\textbf{Performance on the Function, Series and Transforms tasks with increasing complexity} (number of transforms or additional series/functions). \textit{Top row}: accuracy across all 3 tasks with shaded region displaying 95\% Wilson confidence intervals (each point on the plot represents 151 questions). \textit{Bottom row}: accuracy per task decomposition.}
\label{fig:complexity}
\end{figure*}

The \textit{Functions}, \textit{Series} and \textit{Transforms} tasks were designed to include questions spanning a controllable complexity scale. For \textit{Functions} and \textit{Series}, complexity refers to the number of additional functions or series present on the graph, while for \textit{Transforms}, it represents the number of transformations applied to the function. In both cases, the complexity score ranges from 0 (single function/series and no transformations) to 9. In Fig.~\ref{fig:complexity}, we present performance decomposed along this complexity scale for a representative selection of models. Intuitively, we expect performance to decrease as complexity increases due to the added challenge of determining the quantity of interest from additional functions/series or after additional transforms. However, the complexity plots for the evaluated models follow two distinct patterns. For the better performing models, such as Claude 3.5 Sonnet and Gemini 1.5 Pro, performance does clearly decrease as complexity increases from 0 to 3, before remaining around 10\%. For the weaker models, however, the results fluctuate around 10\% across the entire complexity domain. In these cases, even the lowest complexity questions are too challenging. As expected, the questions requiring decimal precision proved much harder than integer precision questions.

\subsection{Evaluation protocols}

\begin{figure*}[t]
\centering
\includegraphics[width=\textwidth]{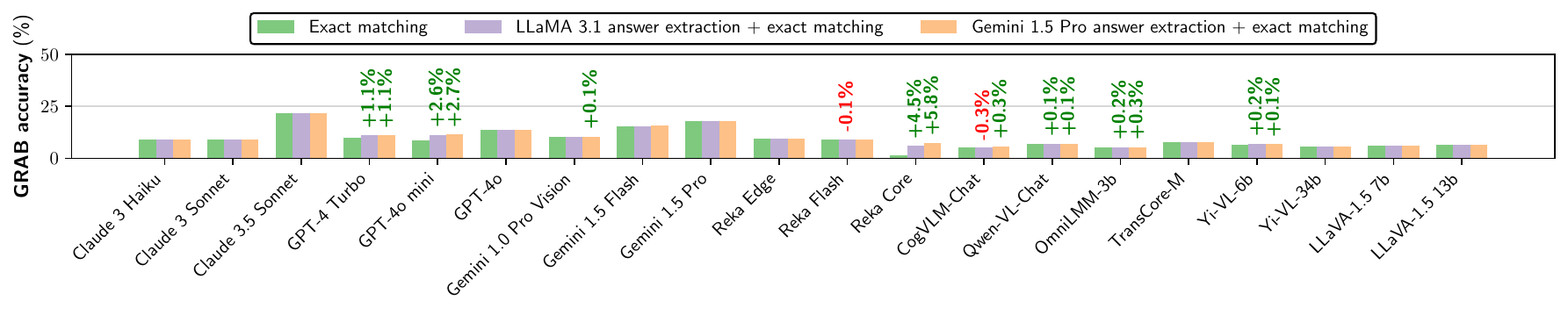}
\vspace{-0.7cm}
\caption{\textbf{Overall accuracy on the \textit{Properties}, \textit{Functions}, \textit{Series} and \textit{Transforms} tasks with different automatic evaluation protocols.} Scores are shown for \textit{exact matching} and \textit{answer extraction (with different LLMs) and then exact matching}. All accuracy differences relative to the exact matching protocol are annotated -- in nearly every case, there is either minor or no difference in score.}

\label{fig:evaluators}
\end{figure*}

In addition to the exact matching evaluation protocol (\S\ref{sec:evaluation}), we explore model performance with more lenient automatic evaluation protocols. Specifically, we utilised an LLM to parse the generated output from the LMMs and extract the answer text, before evaluating it via exact matching as before. We carried this out %
using both a frontier LLM (Gemini 1.5 Pro) and a small open-source LLM (LLaMA 3.1 8b-instruct \cite{llama3.1}). Details of the prompts used can be found in the Appendix. %
We found the LLMs could extract specific answers from verbose output with high accuracy but performed at determining the correctness of a model response given the ground truth answer. A comparison of the different evaluation protocols can be found in Fig.~\ref{fig:evaluators}. \textbf{On the whole, there is almost no difference between the exact matching and automatic evaluation protocols}. Similarly, there is little difference between extraction with a frontier closed-source and a small open-source model, suggesting this approach can feasibly be conducted with little compute. However, \textbf{we advocate for the adoption of exact matching without the use of an LLM as the standard evaluation approach going forwards} as most models are clearly able to sufficiently follow output instructions and it provides a better signal of model capabilities. Cases where there are differences between the performance protocols indicate models that are the poorest instruction followers.

\subsection{Additional ablations}

Additional ablations and analyses can be found in the Appendix, including a comparison of the Matplotlib and Seaborn plotting libraries, alternative evaluation metrics, results on an OCR-based question set, and an analysis of performance with different question formats.

\subsection{GRAB-Lite}

\begin{table}[t]
\resizebox{1\columnwidth}{!}{
\centering
\begin{tabular}{lllllll}
\hline
                & \multicolumn{6}{c}{\textbf{Accuracy (\%)}}               \\
                & \textit{Properties} & \textit{Functions} & \textit{Series} & \textit{Transforms} & \textit{Real} & \textbf{Overall}\\
\textbf{Model}  &        (100)        &           (100)         & (100)          & (100)     & (100)  & (500) \\
\hline
GPT-4o \cite{gpt4o}                       & 21.0 & 7.0 & 10.0 & 6.0 & 19.0 & 12.6\\
Claude 3.5 Sonnet \cite{claude3.5}            & 39.0 & 15.0 & 11.0 & 13.0 & 20.0 & 19.6 \\
Gemini 1.5 Pro \cite{reid2024gemini}              & 23.0 & 10.0 & 14.0 & 7.0 & 25.0 & 15.8\\
\arrayrulecolor[gray]{\linebrightness} \hline
GPT-5 \cite{gpt5}                        & \cellcolor{table_shade}\textbf{59.0} & 34.0 & 33.0 & \cellcolor{table_shade}\textbf{63.0} & \cellcolor{table_shade}\textbf{42.0} & \cellcolor{table_shade}\textbf{46.2} \\
GPT-5 mini \cite{gpt5}                      & 55.0 & \cellcolor{table_shade}\textbf{40.0} & 32.0 & 56.0 & 33.0 & 43.2 \\
GPT-5 nano \cite{gpt5}                      & 36.0 & 34.0 & 29.0 & 33.0 & 19.0 & 30.2 \\
GPT-4.1 \cite{gpt41}                     & 31.0 & 21.0 & 30.0 & 29.0 & 24.0 & 27.0 \\
o1 \cite{o1_reasoning}                          & 27.0 & 15.0 & 28.0 & 26.0 & 25.0 & 24.2\\
Claude Sonnet 4.5 \cite{claude45}            & 47.0 & 34.0 & \cellcolor{table_shade}\textbf{39.0} & 48.0 & 29.0 & 39.4 \\
Claude Sonnet 4 \cite{claude4}             & 37.0 & 31.0 & 31.0 & 29.0 & 24.0 & 30.4 \\
Claude 3.7 Sonnet \cite{claude37}            & 36.0 & 13.0 & 13.0 & 11.0 & 10.0 & 16.6 \\
Gemini 2.5 Pro \cite{gemini25pro}               & 54.0 & \cellcolor{table_shade}\textbf{43.0} & 31.0 & 55.0 & 38.0 & 44.2 \\
Gemini 2.5 Flash \cite{gemini25flash}             & 34.0 & 27.0 & 29.0 & 22.0 & 30.0 & 28.4 \\
Gemini 2.5 Flash Lite \cite{gemini25flashlite}       & 18.0 & 5.0 & 11.0 & 18.0 & 10.0 & 12.4 \\
Gemini 2.0 Flash \cite{gemini20flash}            & 41.0 & 25.0 & 18.0 & 37.0 & 28.0 & 29.8 \\
Gemini 2.0 Flash Lite \cite{gemini20flashlite}       & 14.0 & 13.0 & 9.0 & 14.0 & 12.0 & 12.4 \\
Grok 4 \cite{grok4}                       & 23.0 & 15.0 & 28.0 & 22.0 & 20.0 & 21.6 \\
\arrayrulecolor[gray]{0.1} \hline
\end{tabular}}
\vspace{-0.3cm}
\caption{\textbf{Results on GRAB-Lite}, a lightweight 500-question subset of GRAB designed to evaluate costly reasoning models.}
\label{tab:grab_lite}

\end{table}

Recently, there has been a shift towards \textit{reasoning} models that leverage inference-time scaling and spend more effort ``thinking'' \cite{o1_reasoning}. While this strategy can lead to better-reasoned output, it requires more output tokens, which increases the cost of inference significantly and constrains the scale of evaluation. This creates a pressing need for \textit{lightweight} benchmarks. To address this, we introduce a small 500-question version of our GRAB, called GRAB-Lite. GRAB-Lite is an approximately category, complexity, and task-balanced composition of GRAB with synthetic graphs generated as an equal split between Matplotlib and Seaborn. We evaluate 3 strong GRAB baselines along with newer frontier models (many of which are reasoning models) on GRAB-Lite and report the results in Tab. \ref{tab:grab_lite}. The original baselines attain similar overall scores on both GRAB and GRAB-Lite, with Claude 3.5 Sonnet scoring highest on both. The frontier models also perform well, with GPT-5 achieving the highest overall score, approaching 50\%. An interesting observation is the strong performance of the newly evaluated models on the \textit{Transforms} task, which was the lowest scoring for the GRAB baselines. This suggests the crossing of a capability threshold where the models can now perform the abstract graph transformations and interpret the resulting function to a reasonable degree of accuracy. o1 averaged $>$5k completion tokens (\$0.33) per question, highlighting the need for such a lightweight eval.

\section{Conclusions}

We introduce GRAB, a GRaph Analysis Benchmark for LMMs featuring five tasks covering a diverse range of graph configurations and settings. We leverage synthetic data generation to create high-quality, difficult questions that prove challenging for frontier LMMs, with the best-performing model attaining a score of \highscore\%. On GRAB, we evaluate \nlmmsevaluated closed and open-source models, attaining a broad profile of the abilities of the current generation of LMMs. During inference, we use lightweight prompts with clear output format instructions and assess answer correctness by exact string matching. %
This approach jointly assesses the abilities of models to perform both the  visual perception and reasoning in our tasks and reliably follow output instructions. %
We find little difference in performance using exact matching evaluation vs. automatic LLM evaluation, suggesting most current models can sufficiently follow output instructions, validating exact matching as a suitable evaluation protocol. We also curate a lightweight GRAB-Lite question set designed to evaluate reasoning models. Finally, we conduct studies revealing the categories and splits the models struggle with, and the impacts of different question types and plotting libraries. We release our datasets along with generation and evaluation code for the community to use and hope our work encourages research towards the next generations of LMMs.

\section*{Acknowledgements}
This work was supported by the UKRI Centre for Doctoral Training in Application of Artificial Intelligence to the study of Environmental Risks (reference EP/S022961/1), an Isaac Newton Trust grant, a research gift from Google, an EPSRC HPC grant, the Hong Kong Research Grant Council - General Research Fund (Grant No. 17211024), and HKU Seed Fund for Basic Research. Samuel would like to acknowledge the support of Z. Novak and N. Novak in enabling his contribution.

{
    \small
    \bibliographystyle{ieeenat_fullname}
    \bibliography{main}
}

\newpage
\appendix
\section*{Appendix}
\label{sec:appendix}

We structure this Appendix into 9 parts. In the first part (\S\ref{real}), we include curation details for the Screenshots and Noise splits of the \textit{Real} task questions. In the next four parts, we report additional results and ablations, including a question format ablation (\S\ref{question_format}), a comparison of different performance metrics (\S\ref{additional_metrics}), model performance results on an OCR-based graph analysis question set (\S\ref{ocr}), and a graph plotting libraries performance ablation (\S\ref{plotting_libs}) and display example plots with different libraries. Finally, we detail the inference settings used for GRAB-Lite evaluation (\S\ref{grab-lite_inference}), include examples of the prompts used for inference and LLM evaluation (\S\ref{prompts}) along with details of the compute costs of our work (\S\ref{compute}) and the specific API model versions used throughout this evaluation (\S\ref{apis}).

\section{\textit{Real} task curation details}
\label{real}

\subsection{Screenshots}

To construct the Screenshots split, after creating the question graphs, we randomly selected one of the following digital contexts to embed the image:

\begin{itemize}
\item Presentations (Google Slides, PowerPoint)
\item Video Calling (Teams, Zoom, Meet)
\item Image Viewing (Preview)
\item IDE (VSCode)
\item Email
\item Word Processing (Word, Google Docs)
\end{itemize}

We curated a set of different designs for each digital context (screenshot), including different combinations of background applications/OS. To each, we added an arbitrarily coloured box to the screenshots, denoting where the graph would be embedded. Once created, the coloured box was then replaced with the graph and the composite figure saved and used as part of the question.

\subsection{Noise}

To construct the Noise split, after initially creating the graphs, we randomly applied one of the following types of noise to the image:

\begin{itemize}
\item Gaussian noise
\item Salt and pepper noise
\item Brightness/contrast
\item Blur
\item Spatter/smear
\item JPEG artifacts
\item Rotations
\item Flips
\end{itemize}

\section{Question format ablation}
\label{question_format}

\begin{table}[ht]
\centering
\resizebox{1\columnwidth}{!}{
\centering
\small
\begin{tabular}{llll}
\toprule
& \multicolumn{3}{c}{\textbf{Accuracy (\%)}} \\
\textbf{Model} & Single-answer && Multiple-choice   \\
\midrule
Random chance & - & - & 20.0 \\
\arrayrulecolor{gray!30}\hline\arrayrulecolor{black}
\multicolumn{4}{c}{\light{Closed-source LMMs}} \\
Claude 3 Haiku   \cite{claude3}               & 14.2 & \textcolor{tabg}{+9.4}  & 23.6 \\
Claude 3 Sonnet \cite{claude3}                & 15.3 & \textcolor{tabg}{+10.5} & 25.8 \\
Claude 3.5 Sonnet \cite{claude3.5}            & \cellcolor{table_shade}\textbf{41.8} & \textcolor{tabr}{-10.0}   & 31.8 \\
GPT-4 Turbo \cite{gpt4v}                      & 18.5 & \textcolor{tabg}{+11.3} & 29.8 \\
GPT-4o mini \cite{gpt4o_mini}                 & 15.8 & \textcolor{tabg}{+12.2} & 28.0 \\
GPT-4o \cite{gpt4o}                           & 24.7 & \textcolor{tabg}{+3.0}  & 27.7 \\
Gemini 1.0 Pro Vision   \cite{team2023gemini} & 20.2 & \textcolor{tabr}{-7.5}  & 12.7 \\
Gemini 1.5 Flash \cite{reid2024gemini}        & 28.5 & \textcolor{tabg}{+2.4}  & 30.9 \\
Gemini 1.5 Pro \cite{reid2024gemini}          & 34.2 & \textcolor{tabr}{-1.9}  & \cellcolor{table_shade}\textbf{32.3} \\
Reka Edge \cite{ormazabal2024reka}            & 11.8 & \textcolor{tabg}{+11.5} & 23.3 \\
Reka Flash \cite{ormazabal2024reka}           & 13.2 & \textcolor{tabg}{+7.6}  & 20.8 \\
Reka Core \cite{ormazabal2024reka}            & 1.7  & \textcolor{tabg}{+5.6}  & 7.3  \\
\arrayrulecolor{gray!30}\hline\arrayrulecolor{black}
\multicolumn{4}{c}{\light{Open-source LMMs}} \\
CogVLM-Chat \cite{wang2023cogvlm}             & 7.0  & \textcolor{tabg}{+14.1} & 21.1 \\
Qwen-VL-Chat \cite{bai2023qwen}               & \cellcolor{table_shade}\textbf{10.2} & \textcolor{tabg}{+11.3} & 21.5 \\
OmniLMM-3b \cite{omnilmm}                     & 6.7  & \textcolor{tabg}{+13.6} & 20.3 \\
TransCore-M \cite{TransCore_M}                & 7.9  & \textcolor{tabg}{+12.9} & 20.8 \\
Yi-VL-6b \cite{yi}                            & 5.6  & \textcolor{tabr}{-1.5}  & 4.1  \\
Yi-VL-34b \cite{yi}                           & 7.6  & \textcolor{tabg}{+5.6}  & 13.2 \\
LLaVA-1.5 7b \cite{liu2024improved}  & 4.7 & \textcolor{tabg}{+14.2} & 18.9 \\
LLaVA-1.5 13b   \cite{liu2024improved} & 5.0 & \textcolor{tabg}{+17.4} & \cellcolor{table_shade}\textbf{22.4} \\
\bottomrule
\end{tabular}}
\vspace{-0.3cm}
\caption{\textbf{Accuracy on the \textit{Properties} task with different question formats}. Score differences between the formats are shown as coloured text. The highest open and closed-source model scores for each format are \colorbox{table_shade}{highlighted} and \textbf{bold}.}
\label{tab:mcq}
\end{table}

We carry out an ablation analysing the effect of question format on model performance by re-evaluating the questions from the \textit{Properties} task in a multiple-choice setting with 5 candidate answers. The results of this comparison are displayed in Tab.~\ref{tab:mcq}. Multiple-choice options were generated adversarially by sampling values close to the correct answer. 

For the majority of models, accuracy scores are higher in the multiple-choice setting, and most models score above the random chance score. However, the highest attained score is only 32.3\%, further reflecting the difficulty of the GRAB benchmark. For a few models, including the two leading models, the opposite is true. In these cases, it is possible that the presence of plausible incorrect answers (\ie, close to the true answer) can cause confusion and lead the models to make incorrect selections.

\section{Additional metrics}
\label{additional_metrics}

For a broader view, we report some additional metrics in Tab. \ref{tab:metrics}. In addition to the accuracy scores reported in the main paper (pass@1), here we also report pass@5, 5/5 reliability, root mean squared error (RMSE) and mean absolute error (MAE). These additional metrics largely preserve the pass@1 model rankings, though there is some variation. Our core analysis focuses on pass@1 accuracy as each alternative metric has limitations, making use on GRAB unfeasible. While the pass@5 and 5/5 reliability metrics provide insights into model performance outside of near-deterministic settings, running each evaluation on \nquestions questions five times is impractical. Although most GRAB answers are numeric, distance-based error metrics, such as RMSE and MAE are problematic due to differing scales, ground-truth equal to 0 and treating non-numeric LMM answers.

\begin{table}[h]
\resizebox{1\columnwidth}{!}{
\centering
\begin{tabular}{llllll}
\hline
& \multicolumn{5}{c}{\textbf{Performance metric}}\\
                & \textit{pass@1 $\uparrow$} & \textit{pass@5 $\uparrow$} & \textit{5/5 $\uparrow$} & \textit{RMSE $\downarrow$} & \textit{MAE $\downarrow$}\\
\hline
Claude 3.5 Sonnet & 18.6 & 22.7 & 14.0 & 297.5 & 29.7\\
Gemini 1.5 Pro & \cellcolor{table_shade}\textbf{21.9} & \cellcolor{table_shade}\textbf{28.9} & \cellcolor{table_shade}\textbf{14.7} & \cellcolor{table_shade}\textbf{209.9} & 28.3\\
GPT-4o & 18.0 & 26.0 & 8.3 & 226.9 & \cellcolor{table_shade}\textbf{27.7}\\

\hline
\end{tabular}}
\vspace{-0.2cm}
\caption{\textbf{Performance metrics on \textit{Real} Screenshots split.}}
\label{tab:metrics}
\end{table}

\section{OCR experiments}
\label{ocr}

We construct a small set of 35 OCR questions based on text in the legend, title, and axis labels. The graphs are created using a similar pipeline to the \textit{Properties} task except rather than using purely randomly selected data generation processes, we use plausible data ranges and functions for the axis label. We leverage GPT-4 Turbo to construct a database of `realistic' dependent and independent variables along with their ranges, distributions and directions (\eg linear, increasing). We evaluate these questions in a multiple-choice setting with 5 options and present the results in Tab~\ref{tab:ocr}. Compared to performance on the Properties task with multiple-choice options (Tab.~\ref{tab:mcq}), the models attain much higher scores on these OCR-style questions, with many models achieving either 100\% or close to 100\% accuracy. These high scores suggest this particular question type is too easy for current frontier LMMs, therefore we refrain from including it in GRAB.

\begin{table}[htb]
\centering
\begin{tabular}{ll}
\toprule
\textbf{Model} & \textbf{Accuracy (\%)} \\
\midrule
Random chance         & 20.0 \\
\arrayrulecolor{gray!30}\hline\arrayrulecolor{black}
\multicolumn{2}{c}{\light{Closed-source LMMs}} \\
Claude 3 Haiku \cite{claude3}        & 42.9  \\
Claude 3 Sonnet \cite{claude3}       & 91.4  \\
Gemini 1.0 Pro Vision \cite{team2023gemini} & \cellcolor{table_shade}\textbf{100.0} \\
Gemini 1.5 Flash \cite{reid2024gemini}      & \cellcolor{table_shade}\textbf{100.0} \\
Gemini 1.5 Pro \cite{reid2024gemini}        & \cellcolor{table_shade}\textbf{100.0} \\
GPT-4 Turbo \cite{gpt4v}           & 97.1  \\
GPT-4V \cite{gpt4v}                & 97.1  \\
GPT-4o \cite{gpt4o}                & 82.9  \\
Reka Core \cite{ormazabal2024reka}             & 97.1  \\
Reka Edge \cite{ormazabal2024reka}             & 88.6  \\
Reka Flash \cite{ormazabal2024reka}            & \cellcolor{table_shade}\textbf{100.0} \\
\arrayrulecolor{gray!30}\hline\arrayrulecolor{black}
\multicolumn{2}{c}{\light{Open-source LMMs}} \\
TransCore-M \cite{TransCore_M}           & \cellcolor{table_shade}\textbf{80.0}  \\
Yi-VL-6b \cite{yi}              & 51.4  \\
OmniLMM-3b \cite{omnilmm}            & 65.7  \\
Qwen-VL-Chat \cite{bai2023qwen}          & 71.4  \\
\bottomrule
\end{tabular}
\vspace{-0.3cm}
\caption{\textbf{Accuracy scores on OCR experiments}.}
\label{tab:ocr}
\end{table}

\section{Plotting libraries ablation}
\label{plotting_libs}

\begin{figure*}[t]
    \centering
    \begin{subfigure}{0.45\textwidth}
        \centering
        \includegraphics[width=\textwidth]{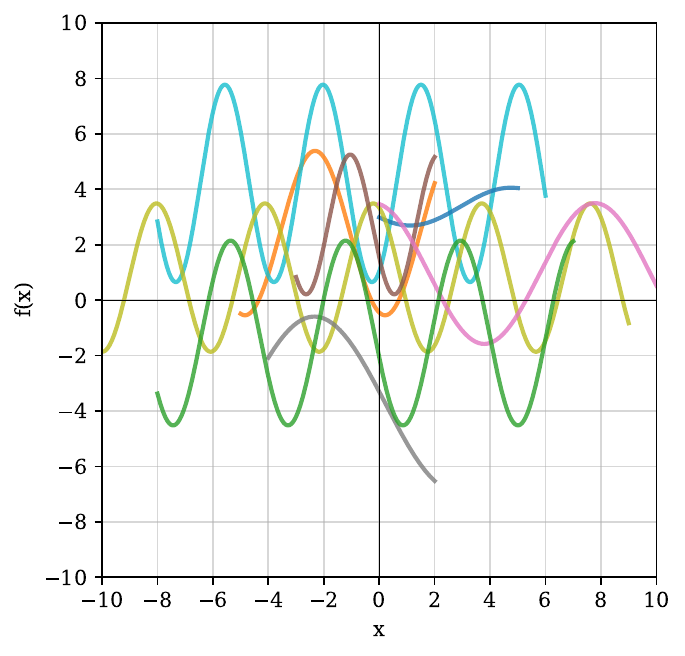}
        \caption{Matplotlib \textit{Functions} example}
        \label{fig:sub1}
    \end{subfigure}
    \hfill
    \begin{subfigure}{0.45\textwidth}
        \centering
        \includegraphics[width=\textwidth]{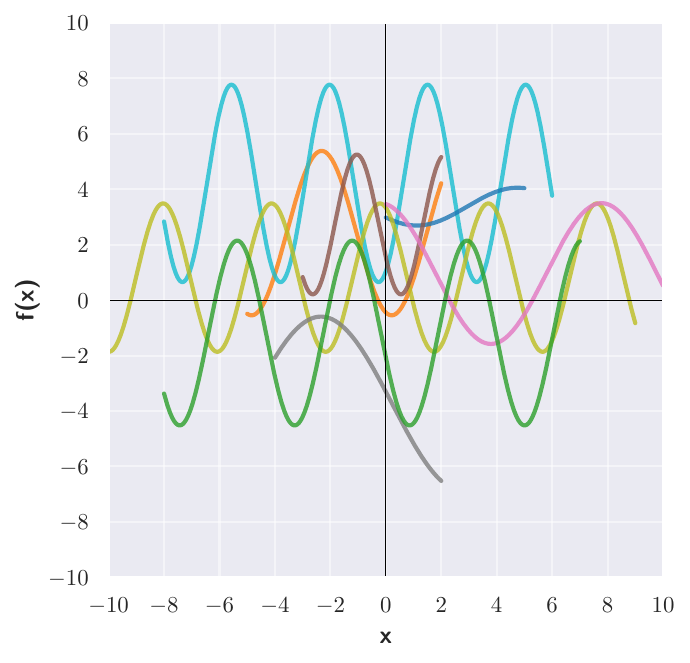}
        \caption{Seaborn \textit{Functions} example}
        \label{fig:sub2}
    \end{subfigure}

    \vspace{1em}  %

    \begin{subfigure}{0.45\textwidth}
        \centering
        \includegraphics[width=\textwidth]{arxiv/figures/_amplitude_8_50_plt.pdf}
        \caption{Matplotlib \textit{Series} example}
        \label{fig:sub3}
    \end{subfigure}
    \hfill
    \begin{subfigure}{0.45\textwidth}
        \centering
        \includegraphics[width=\textwidth]{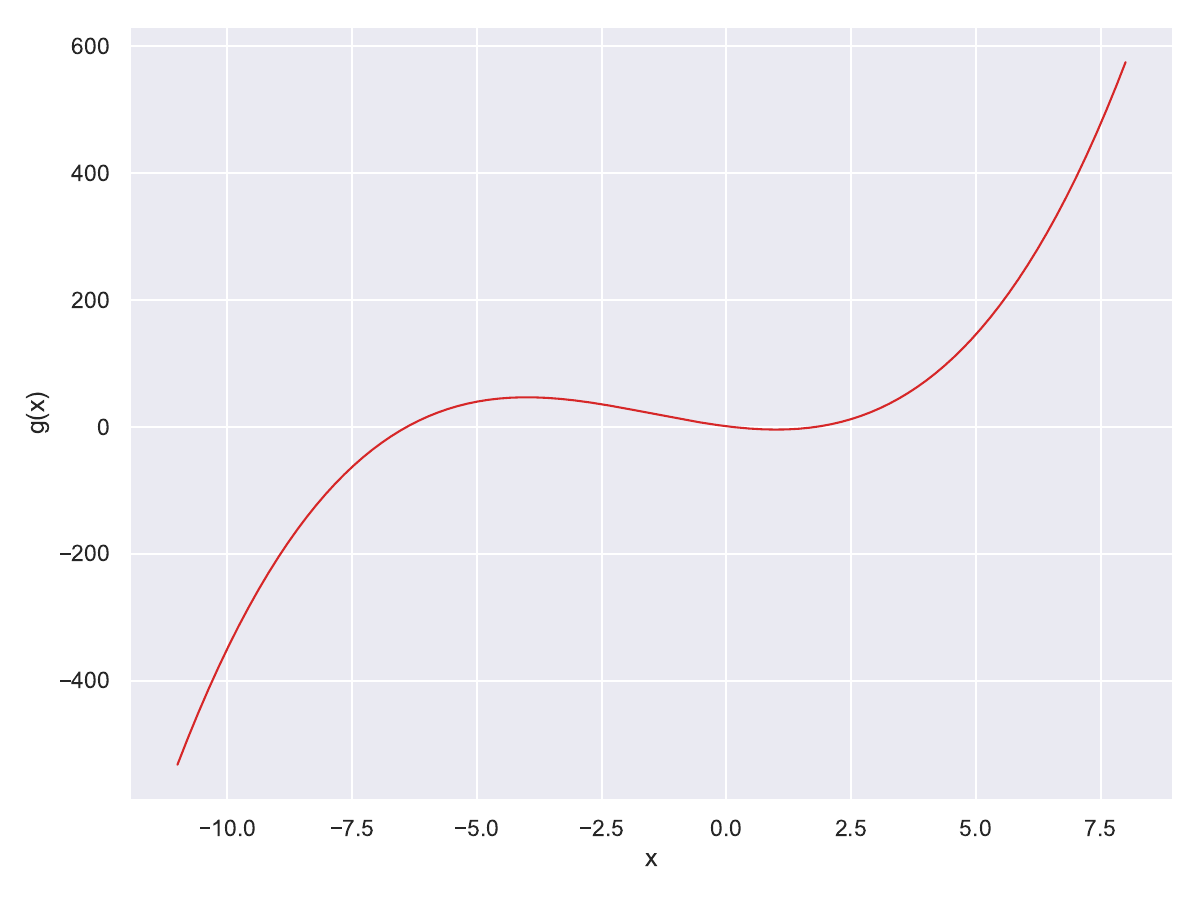}
        \caption{Seaborn \textit{Series} example}
        \label{fig:sub4}
    \end{subfigure}
    \caption{Example equivalent plots from the Matplotlib and Seaborn plotting libraries.}
    \label{fig:2x2grid}
\end{figure*}

To compare different plotting libraries, we sample 100 GRAB questions and synthesise them using both Matplotlib \cite{plt} and the Seaborn \cite{Waskom2021} library to create two sets of questions that differ in the aesthetics and styles of the libraries but are otherwise identical. The near agreement of Claude 3.5 Sonnet's scores on these sets (Tab. \ref{tab:plotting_libraries}) suggests that plotting with different libraries does not significantly impact overall performance; therefore, we focus on Matplotlib plots for GRAB. However, to increase the diversity of GRAB-Lite, we re-generate half of the synthetic plots using Seaborn. Fig. \ref{fig:2x2grid} displays examples of identical functions and series plotted with both the Matplotlib and Seaborn libraries.

\begin{table}[h]
\resizebox{1\columnwidth}{!}{
\centering
\begin{tabular}{llllll}
\hline
& \multicolumn{5}{c}{\textbf{Accuracy (\%)}}\\
                & \textit{Properties} & \textit{Functions} & \textit{Series} & \textit{Transforms} & \textbf{Overall}\\
\textbf{Plotting Library}  &        (25)        &           (25)         & (25)          & (25)       & (100) \\
\hline
Matplotlib \cite{plt}      & \cellcolor{table_shade}\textbf{48.0} & 16.0 & \cellcolor{table_shade}\textbf{36.0} & \cellcolor{table_shade}\textbf{20.0} & \cellcolor{table_shade}\textbf{30.0} \\
Seaborn \cite{Waskom2021}                & 40.0 & \cellcolor{table_shade}\textbf{20.0} & 24.0 & \cellcolor{table_shade}\textbf{20.0} & 26.0 \\

\hline
\end{tabular}}
\vspace{-0.3cm}
\caption{\textbf{Claude 3.5 Sonnet accuracy on different plotting libraries} on a 100-question subset of GRAB.}

\label{tab:plotting_libraries}
\end{table}

\section{GRAB-Lite inference}
\label{grab-lite_inference}

For inference on the GRAB-Lite questions we evaluated o1, GPT4.1 and the GPT-5 models via the OpenAI API \cite{openaiapi}. The Gemini Developer API \cite{geminiapi} was used to evaluate Gemini 2.0 Flash and Flash Lite, Gemini 2.5 Flash and Flash Lite, and Gemini 2.5 Pro. We used the Vertex AI API \cite{vertexaiapi} to evaluate Claude 3.7 Sonnet and Claude Sonnet 4 and 4.5, and the xAI API \cite{xaiapi} for Grok 4. Greedy decoding was used where possible. The `reasoning effort' parameter for o1, GPT-5 mini and nano was set to `high'; for GPT-5 it was set to `medium'.

\section{Prompts}
\label{prompts}

Inference prompt:
\begin{formattedquote}
    \textit{<question>\textbackslash n Only provide the answer, no reasoning steps.
    If you are unsure, still provide an answer. Answer:\textbackslash n '}
\end{formattedquote}

\noindent LLM evaluation prompt:
 \begin{formattedquote}
    \textit{A generative model has been asked this question: "<question>" about a plot.\textbackslash n
    The output from the model answering the question is: "<output>"\textbackslash n
    Extract just the answer from the generative model output. Maintain the same precision given by the model. 
    Convert any numbers to digits (e.g., "one" to "1"). Remove any additional terms like 'approximately'.
    Return only the extracted numeric answer, without any additional text or explanation. If no answer is provided, return "None".}
\end{formattedquote}

\section{Compute}
\label{compute}

As we restricted the size of our benchmark to \nquestions, the total compute required for this work is relatively low. Negligible compute was needed to create the synthetic images. Inference with the closed-source models was carried out via API calls. Using a single NVIDIA A100-80GB GPU, inference on the entire GRAB benchmark using LLaVA-1.5 7b can be carried out in approximately 45 minutes, using a single process (inference time is significantly reduced with multiprocessing).

\section{API model versions}
\label{apis}
These are the specific versions of the API models used in this work:

\begin{itemize}
    \item GPT-4 Turbo:\\\textit{gpt-4-turbo-2024-04-09}
    \item GPT-4o mini:\\\textit{gpt-4o-mini-2024-07-18}
    \item GPT-4o:\\\textit{gpt-4o-2024-05-13}
    \item GPT-4.1:\\\textit{gpt-4.1-2025-04-14}
    \item o1:\\\textit{o1-2024-12-17}
    \item GPT-5 nano:\\\textit{gpt-5-nano-2025-08-07}
    \item GPT-5 mini:\\\textit{gpt-5-mini-2025-08-07}
    \item GPT-5:\\\textit{gpt-5-2025-08-07}
    \item Gemini Pro Vision:\\\textit{gemini-1.0-pro-vision-001}
    \item Gemini 1.5 Flash:\\\textit{gemini-1.5-flash-preview-0514}
    \item Gemini 1.5 Pro:\\\textit{gemini-1.5-pro-preview-0514}
    \item Gemini 2.0 Flash:\\\textit{gemini-2.0-flash-001}
    \item Gemini 2.0 Flash Lite:\\\textit{gemini-2.0-flash-lite-001}
    \item Gemini 2.5 Flash:\\\textit{gemini-2.5-flash-preview-05-20}
    \item Gemini 2.5 Flash Lite:\\\textit{gemini-2.5-flash-lite}
    \item Gemini 2.5 Pro:\\\textit{gemini-2.5-pro}
    \item Claude 3 Haiku:\\\textit{claude-3-haiku@20240307}
    \item Claude 3 Sonnet:\\\textit{claude-3-sonnet@20240229}
    \item Claude 3.5 Sonnet:\\\textit{claude-3-5-sonnet@20240620}
    \item Claude 3.7 Sonnet:\\\textit{claude-3-7-sonnet@20250219}
    \item Claude Sonnet 4:\\\textit{claude-sonnet-4@20250514}
    \item Claude Sonnet 4.5:\\\textit{claude-sonnet-4-5@20250929}
    \item Reka Edge:\\\textit{reka-edge-20240208}
    \item Reka Flash:\\\textit{reka-flash-20240226}
    \item Reka Core:\\\textit{reka-core-20240501}
    \item Grok-4:\\\textit{grok-4-0709}

\end{itemize}

\end{document}